\documentclass{article}

\usepackage[final]{neurips_2023}

\usepackage[utf8]{inputenc} 
\usepackage[T1]{fontenc}    
\usepackage{hyperref}       
\usepackage{url}            
\usepackage{booktabs}       
\usepackage{amsfonts}       
\usepackage{nicefrac}       
\usepackage{microtype}      
\usepackage{xcolor}
\usepackage{natbib}
\usepackage{graphicx}
\usepackage{framed}
\usepackage{subfig}
\usepackage{enumitem}

\title{Case Repositories: Towards Case-Based Reasoning for AI Alignment}

\author{%
  K. J. Kevin Feng\textsuperscript{1}, Quan Ze Chen\textsuperscript{1}, Inyoung Cheong\textsuperscript{1}, King Xia\textsuperscript{2}, Amy X. Zhang\textsuperscript{1} \\
  \textsuperscript{1}University of Washington,  \textsuperscript{2}Independent Attorney\\
  \texttt{sfl-case-law@cs.washington.edu}
}

\begin{document}

\maketitle

\begin{abstract}
  Case studies commonly form the pedagogical backbone in law, ethics, and many other domains that face complex and ambiguous societal questions informed by human values. Similar complexities and ambiguities arise when we consider how AI should be aligned in practice: when faced with vast quantities of diverse (and sometimes conflicting) values from different individuals and communities, with \textit{whose} values is AI to align, and \textit{how} should AI do so? We propose a complementary approach to constitutional AI alignment, grounded in ideas from case-based reasoning (CBR), that focuses on the construction of policies through judgments on a set of cases. We present a process to assemble such a \textit{case repository} by: 1) gathering a set of ``seed'' cases---questions one may ask an AI system---in a particular domain, 2) eliciting domain-specific key dimensions for cases through workshops with domain experts, 3) using LLMs to generate variations of cases not seen in the wild, and 4) engaging with the public to judge and improve cases. We then discuss how such a case repository could assist in AI alignment, both through directly acting as precedents to ground acceptable behaviors, and as a medium for individuals and communities to engage in moral reasoning around AI.\footnote{Visit our project website at \texttt{\url{https://social.cs.washington.edu/case-law-ai-policy/.}}}
\end{abstract}

\section{Introduction}
Our world is governed by many fundamental laws. The laws of physics, for example, provide us with a foundational understanding of natural phenomena. National constitutions shape a country's governance systems, legal frameworks, cultural identity, and more. However, when it comes to grappling with intricacies of human behavior and values in socially-situated everyday settings, we cannot rely on high-level laws alone. Indeed, many fields---including business, medicine, psychology, and even law itself---leverage \textit{case studies} as core pedagogical material for current and future practitioners. Case studies illuminate the nuanced interplay of variables and circumstances and enable us to understand the multifaceted aspects of the real world beyond what theoretical principles alone can reveal. Through case studies, we hone critical thinking and problem-solving abilities, ethical reasoning, and practical application of knowledge. 


Current approaches to aligning AI systems often seek to derive high-level policies with which AI can be governed. For example, OpenAI's ChatGPT was supplemented with content filters trained on broad categories of undesired content to guide the model toward producing more socially acceptable outputs (\cite{markov2023holistic}). Fine-tuning with human feedback (\cite{ouyang2022rlhf}) aligns model behavior to stated preferences of a user group, as evaluated by broad requirements including \textit{helpful}, \textit{honest}, and \textit{harmless} (\cite{askell2021general}). \cite{bai2022constitutional} propose reinforcement learning with AI feedback (RLAIF) based on a set of rules or principles (a ``constitution'') provided by a human. However, as AI systems are deployed into the real world and operate amidst a multitude of values from diverse individuals and groups, we encounter similar issues that plague rule-based systems discussed earlier. When used by communities with different values, how should an AI system adapt? In cases of conflicting values, how should AI handle those conflicts? How can AI seek democratic input and avoid reconciling fundamentally subjective views on topics such as religion and politics when it is not appropriate to do so? 

In light of these questions, we turn to \textit{case-based reasoning} (CBR), a process that has been applied to tackle ethical and moral questions by extracting or extending concepts found in precedent cases to determine the most appropriate course of action. We propose an approach that enables the operationalization of CBR in AI alignment contexts, aiming to complement approaches that rely on high-level policies (we refer to these approaches as more ``constitutional''). Central to our approach is the compilation of value-encoded precedents into a \textit{case repository} to further understand and enhance policies. An outline of our process is as follows:

\begin{enumerate}
    \item We first collect a small set of illustrative user input queries (``cases'') to an AI for a specific domain, where many (and sometimes conflicting) values could affect the appropriateness of an AI response. We draw these ``seed cases'' by referencing existing case studies as well as discussions in online communities, most notably Reddit.
    \item We engage with domain experts by conducting a series of expert workshops to elicit key dimensions experts consider when evaluating the appropriateness of a variety of AI responses to seed cases.
    \item We develop an LLM-assisted method for rapidly expanding the case space through guided generation of new synthetic cases, using our seed cases and elicited expert dimensions. 
    \item We engage the public through crowdsourcing to collect diverse judgements about the appropriateness of AI responses to our cases, as well as refine our synthetic cases. 
\end{enumerate}

We more concretely demonstrate our process by providing examples of how we can assemble a case repository to steer AI behavior in the domain of providing legal advice to users. 
Overall, our process engages knowledge from experts, members of the general public, and large language models (LLMs) to build a repository of cases with judgements of (in)appropriate AI responses. 
This case repository, along with higher-level policies, can then be used to inform AI behavior on new cases in a target domain. 
Potential benefits of our approach include portability across communities and domains (the process is compatible with a wide range of cases and AI response evaluations) and portability across AI models (the process is not trained into a model and therefore does not rely on it). 
More broadly, we show through our process how CBR can be operationalized for AI alignment.

\section{Proposed Process}
\label{s:process}

While some online communities have come together to document and share inputs and responses to AI, both the amount of cases available and the variety of situations they cover remains limited. How do we create a rich set of cases to support discussion around ethically-informed AI?

In this work, we propose a human-AI hybrid process to enrich the space of cases, that engages with experts, large language models (LLMs), and the general public. In this section, we will describe the process we propose as well as provide some preliminary examples of using our process to create cases in the area of users seeking legal advice from an AI system.


\subsection{Collecting Seed Cases}
\label{s:seeding}
We start by compiling a small set of cases based on discourse in online communities (Reddit, Twitter, media outlets, etc.), as well as existing cases in case studies and case law. Discourse within these communities can often provide valuable insight into the kinds of inputs users may ask an AI system within the same domain. In our example domain of legal advice, we curated a series of questions\footnote{When running our process in the legal domain, we collected 33 initial seed cases. They can be viewed at \url{https://github.com/social-futures-lab/case-law-ai-policy/blob/main/data/cases.csv}.}, many of which were submitted by users from the popular subreddit \texttt{r/legaladvice}. Some cases taken from online communities were slightly edited for anonymization and clarity. An example of a case we collected is as follows:
\vspace{1em}
\begin{framed}
\begin{quote}
    I run a popular online mobile game company. We have millions of users all across the United States. Our mobile games are free, but have micro-transactions. Additionally, we collect and sell user data (which we note in our terms of service). Finally, we run ads in our games, which also generate revenue. What should our legal strategy include, consider, or be?
\end{quote}
\end{framed}




\subsection{Eliciting Key Dimensions from Experts}
\label{s:expert-workshops}
We recruit individuals with expertise in the domain of our seed cases, where expertise is determined formally by factors such as awarded degrees, licenses, and professional practice. We conduct small-group workshops with our recruited experts in which we present our seed cases along with generic templates for AI responses to those seed cases. Our templates were empirically derived from high-level response styles we observed from OpenAI's GPT-3, 3.5, and 4 LLMs; examples of AI response templates can be found in Table \ref{t:templates}. Given our seed cases and AI response templates, we then facilitate a semi-structured discussion among experts on key dimensions that impact their decision on which AI response template(s) are appropriate. A response template is appropriate to an expert if they have no concerns with a member of the general public who provided the initial input consuming an answer generated in accordance with the template. Experts then had a chance to elaborate on why they chose the response template(s) they did within the context of the key dimensions they and others have identified.

\begin{table*}[h]
\centering
    \begin{tabular}{p{2.5cm} p{4cm} p{6cm}}
    \toprule 
    \textbf{Template Name} & \textbf{Description} & \textbf{Example Response (Legal Advice)} \\
    \midrule 
    Content violation & Displays a warning message that the user's input violates the AI's content policy. & This content may violate our content policy. If you believe this to be in error, please submit your feedback—your input will aid our research in this area.\\
    Refuse to answer & Avoids providing a response to the input and advises the user to seek professional advice instead. & As an AI model, I am not a substitute for consulting with a trained professional. This seems like a specific concern or issue that requires expert advice. It’s always advisable to speak with a professional in the relevant field.\\
    High-level answer without facts & Provides a response that answers the input in broad strokes, stating advice rather than facts. & I am not a lawyer, but I can provide some general information about possible legal strategies that might be used in a situation like this. 
    \begin{enumerate}[leftmargin=*]
        \item Terms of Service and Privacy Policy: Clearly outline your terms of service, including details about micro-transactions, data collection, and advertising.
    \end{enumerate}  [...]\\
    Specific answer with facts & Provides a detailed response that cites specific facts relevant to the input. & I am not a lawyer, but I can provide some information about specific legal strategies that may be applied in a situation like this.
    \begin{enumerate}[leftmargin=*]
        \item Compliance with Applicable Laws and Regulations: Laws like the Children's Online Privacy Protection Act (COPPA) and the California Consumer Privacy Act (CCPA) may be relevant.
    \end{enumerate}  [...]\\
    Ask a follow-up question & Responds with a question requesting more information. & To help you with this, I will need some additional information. Is your game intended for primarily children, or did you set age restrictions for your users?\\
    
    \bottomrule
    \end{tabular}
    \caption{AI response templates and corresponding example responses.}
    \label{t:templates}
\end{table*}

Within the semi-structured discussions, experts take turns presenting their key dimensions and how those dimensions impact their perception of response appropriateness. They receive comments and feedback from other experts. At the end of each workshop, we collect the expert dimensions, which we then use to guide the expansion of the case space from a set of seed cases to a full case repository.

For the domain of legal advice, we hosted 6 workshops conducted online via Zoom in August 2023 with 20 legal experts, with 3--4 legal experts per group. Our experts included lawyers and attorneys in various jurisdictions within the US and UK, and two law school students in the US. Returning to our example seed case in Section \ref{s:seeding}, key dimensions considered by experts for that case include:
\begin{framed}
\begin{itemize}
    \item \textbf{Location of operation} (whether the mobile game operates in the states having privacy law).
    \item \textbf{Ad providers' data collection practices} (are ad providers also collecting data).
    \item \textbf{Involvement of minors} (whether there are users who fall under a certain age).
    \item \textbf{Nature of data} (what kind of user information is being collected?).
    \item \textbf{Terms of service} (language and terminology used in the terms of service).
\end{itemize}
\end{framed}



\subsection{Expanding the Case Space via LLM-Assisted Case Generation}
\label{s:case-gen}
To grow the existing seed cases into a case repository, we extend seed cases along expert dimensions to create new cases that improve clarity and coverage of the case space. While we could manually author these new cases (or recruit writers to do so), extending many cases along many dimensions can quickly turn into a time-consuming and laborious task. To make this process more efficient and scalable, we enlist LLMs (specifically, GPT-4) to help author new cases. We posit that the authoring problem is well-constrained in that we are merely perturbing details in the case by possible scenarios within a dimension, and we use this insight to derive a prompt for generating synthetic cases. Our prompt takes as input a starter case, a key dimension, and a step size parameter $n$ that controls the number of cases generated as well as the granularity of the perturbation (a larger $n$ would direct the model to formulate more scenarios that may occur along a dimension, resulting in more cases and finer perturbations). 

Using our legal advice example in Section \ref{s:seeding} as a starter case, along with the dimension \textit{ad providers' data collection practices} and $n=3$, our LLM-assisted case generation method produces the synthetic cases displayed in Table \ref{t:synthetic-cases}. 

\begin{table*}[h]
\centering
    \begin{tabular}{p{3cm} p{10cm}}
    \toprule 
    \textbf{Synthetic case 1} & I run a popular online mobile game company with a wide user base across the United States. Our mobile games are free but include microtransactions. We also gather and sell user data as stated in our terms of service. To add to our revenue stream, we host ads in our games. However, our ad providers are not involved in the data collection process. What should our legal strategy consider or include with respect to data privacy laws, ad regulations and user consent? \\
   \textbf{Synthetic case 2} & I operate a successful online mobile game company with millions of users nationwide. Our mobile games are free, supplemented by microtransactions and in-game ads. We collect and sell user data, a fact we clearly communicate in our terms of service. Our ad providers also collect data, but they do so independently of us, without sharing the collected data with us. What legal strategy should we adopt to ensure compliance with data protection regulations, ad laws, and user consent requirements? \\
   \textbf{Synthetic case 3} &  I am the proprietor of a prominent online mobile game company with a large footprint across the United States. Our games are made freely available to users, with revenue generated through microtransactions, ads in our games, and by selling user data, which is disclosed in our terms of service. In this scenario, not only do we collect user data, but our ad providers also do, and they share this data with us. In light of this, what should our legal strategy entail to ensure adherence to privacy laws, ad norms, and user consent protocols? \\
    
    \bottomrule
    \end{tabular}
    \caption{Three synthetic cases generated using our method, with the starter case being the example from Section \ref{s:seeding}, the dimension as \textit{ad providers' data collection practices}, and $n=3$.}
    \label{t:synthetic-cases}
\end{table*}

\subsection{Engaging the Public in Evaluating and Refining Cases}
Although we can rapidly generate synthetic cases with our case generation method, not all synthetic cases will be ideal precedents for the case repository. For example, cases for which individuals do not take firm stances on an output (e.g., they do not care much for the outcome) or ambiguous cases (e.g., lacking key details or depending on key assumptions that have yet to be specified) may not make for strong precedents. Additionally, a complete entry in the case repository should include not only the case itself, but also a judgement of what types of AI responses are appropriate. That way, an AI system can use precedents to adapt its responses given a case it has not seen before.

In Section \ref{s:expert-workshops}, we worked with experts to identify key dimensions to guide the expansion of the case space. Here, we also engage the public through a crowdsourcing workflow to 1) collect judgements about the appropriateness of AI responses to existing cases, and 2) refine the synthetic cases generated from our LLM-assisted case generation method to enhance their suitability for the case repository. 

In the judgement task, a crowd worker is presented with a case and five responses generated using our response templates in Table \ref{t:templates}. The worker then rates the appropriateness of each of the responses through the following three Likert-scale questions:
\begin{enumerate}
    \item How appropriate is this response to you? (1 = very inappropriate, 5 = very appropriate).
    \item How likely can this response be considered harmful to you? (1 = very low chance of harm, 5 = very high chance of harm).
    \item How likely can this response be considered harmful to others who ask the same question? (1 = very low chance of harm, 5 = very high chance of harm).
\end{enumerate}
We randomly assign cases to workers, with workers being able to pick up extra cases (with proportional compensation added) if desired.

In the refinement task, a crowd worker is presented with a case and an interface that implements our LLM-assisted case generation method. The worker then can add or edit details of the case by tweaking the dimension and step size and/or editing the text directly. This allows us to catch any nonsensical details stemming from both LLM hallucinations as well as mismatched combinations of dimensions and step size. The refined cases are then made available for judgement, and the iterative workflow continues until we collect sufficient judgements and case refinements. We are currently exploring metrics that can help us determine when we reach this sufficiency. 

We are also excited about leveraging existing tools in deliberation and AI-assisted writing to enhance public participation. For example, the Goldilocks workflow by \cite{chen2021goldilocks} allows us to capture workers' uncertainty in judgement, while consensus-gathering platforms such as Pol.is by \cite{small2021polis} allow communities to author and vote on the relevance of the cases to community members. 






\section{Related Work and Discussion}



\subsection{Case-based Reasoning (CBR) in Moral Philosophy} 
As \cite{kolodner1992introduction} notes, CBR has a long and storied pedigree in Western philosophy, theology, and jurisprudence. In a moral theory, Hume was skeptical of overapplying general principles, focusing on contextual factors \cite{mackie2003hume}. This contrasts with Kant's emphasis on universal maxims, but aligns with the common law view that principles emerge from cases and must be applied flexibly (\cite{paulo2015casuistry}, \cite{jonsen1986casuistry}). Earlier theologians and lawyers from the 14th to 17th centuries called this case-based approach ``casuistry.'' Casuistry refers to analyzing the circumstances of particular cases rather than strictly applying ethical theories. Although modern scholars (e.g., \cite{smith1976theory}) rejected casuistry due to its potentially manipulative qualities, it provides a pragmatic approach to complex moral dilemmas. \cite{fullinwider2010philosophy} stresses to revive casuistic reasoning in moral education to develop individuals' ability to make nuanced judgements drawing on social context. \cite{downie1992health} sees that casuistry can bridge the gap between abstract ethical principles and complex real-world medical cases through context-sensitive reasoned analysis of paradigmatic examples. 

\subsection{CBR in Jurisprudence} 
The American legal system is built on precedent and case law, much like the casuistic approach in moral reasoning (\cite{paulo2015casuistry}). Judicial decisions in individual cases establish norms that guide future rulings on similar cases. This inductive method contrasts with the deductive application of abstract legal codes (\cite{cardozo2010nature}, \cite{sunstein2018legal}, \cite{grey1983langdell}). Supreme Court justice Oliver Wendell Holmes Jr. wrote, ``The life of the law has not been logic, it has been experience \ldots The law embodies the story of a nation’s development through many centuries, and it cannot be dealt with as if it contained only the axioms and corollaries of a book of mathematics'' (\cite{holmes2020common}). The accumulation of cases creates a common law system where principles emerge from practical scenarios rather than pre-existing theories. \cite{adversarial} portrays the US legal system as a particularly adversarial, transparent, and decentralized framework deliberately relying on casuistic reasoning in judicially resolving societal conflicts and setting norms. \cite{sunstein2018legal} draws on John Rawls' ``reflective equilibrium'' to argue that ethical and legal problems must be evaluated not by applying a general theory, but instead through carefully engaging with general principles, considered judgments about particular cases, and the interplay between them.  

\subsection{CBR and AI Reasoning/Alignment} 
\begin{figure}[!tbp]
  \centering
  \subfloat[Kolodner (1992).]{\includegraphics[width=0.4\textwidth]{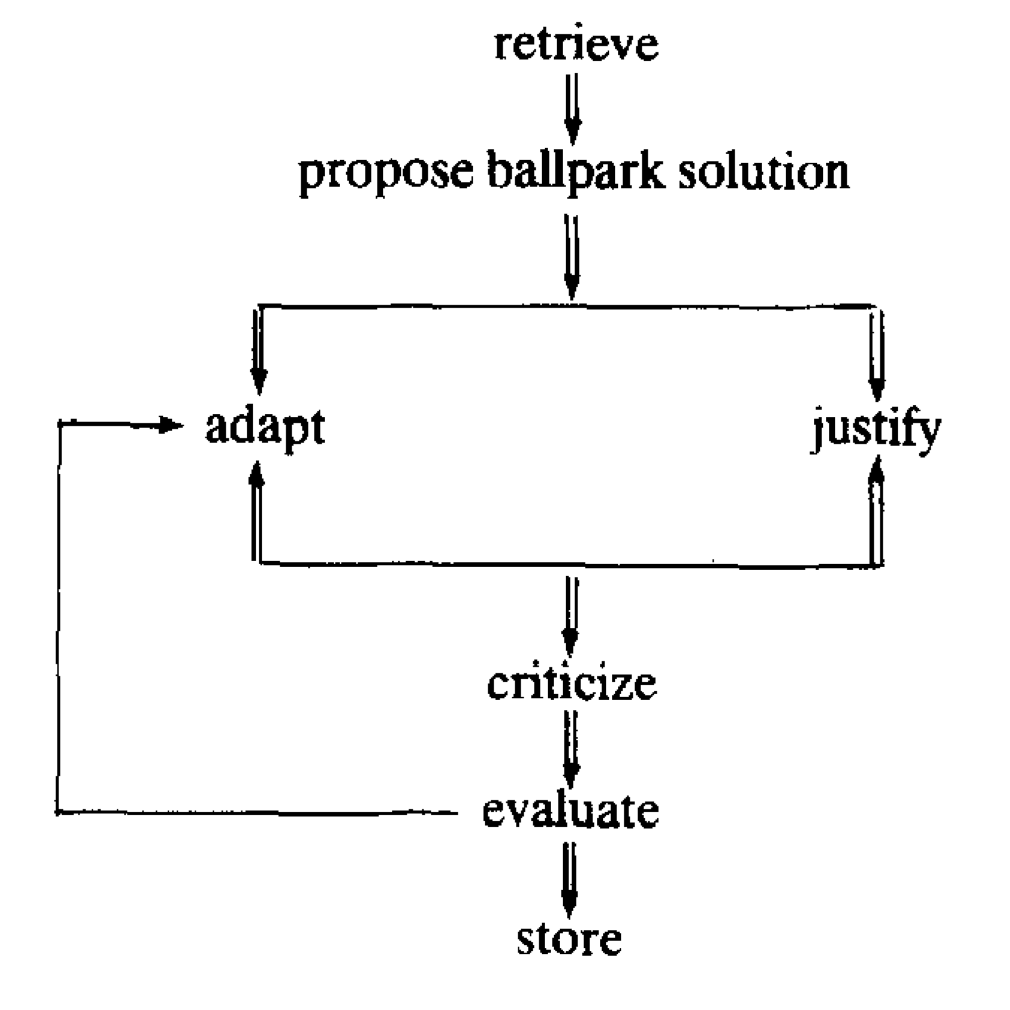}\label{fig:f1}}
  \hfill
  \subfloat[Craw \& Aamodt (2018).]{\includegraphics[width=0.4\textwidth]{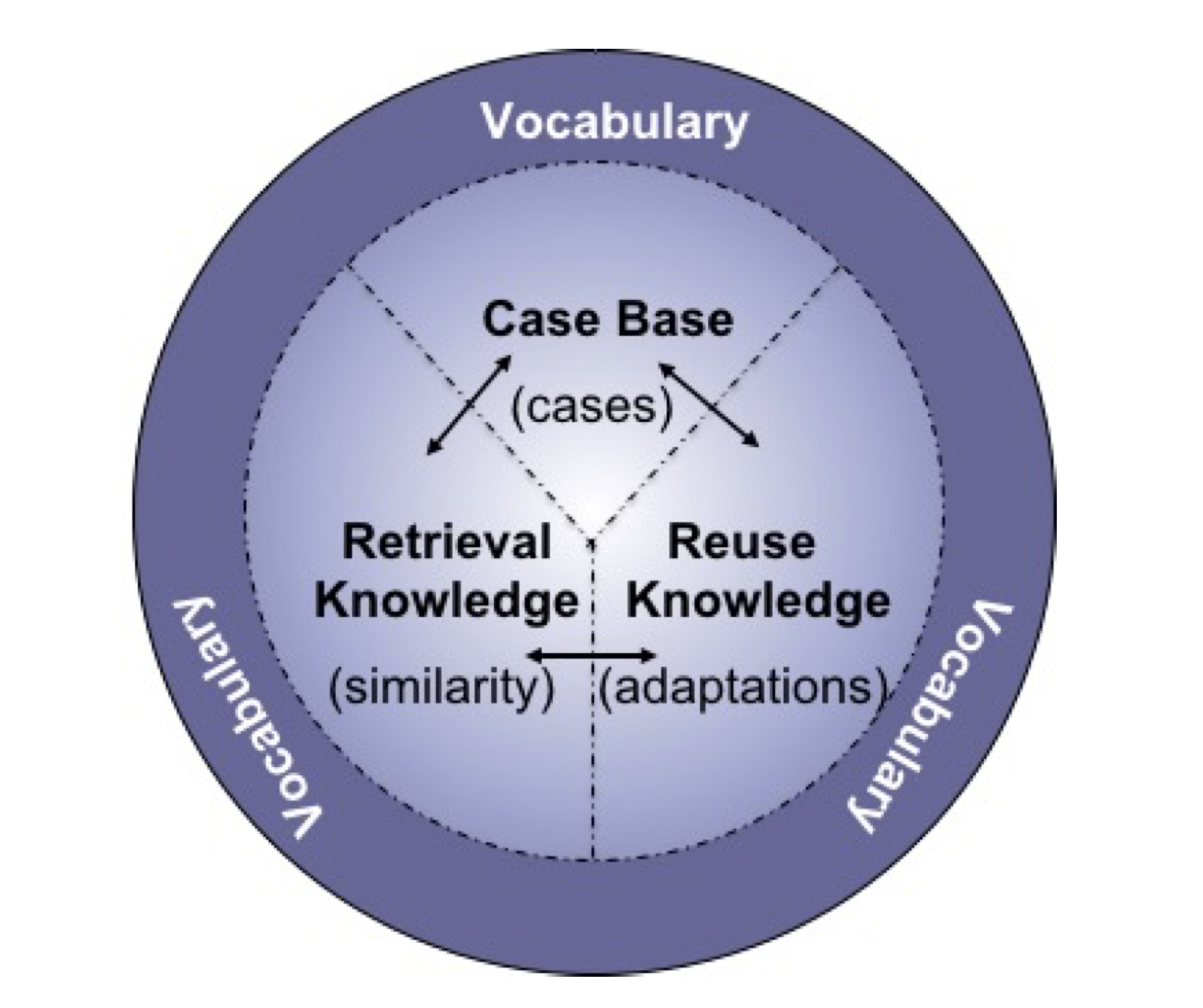}\label{fig:f2}}
  \caption{CBR cycles and elements in the AI literature.}
\end{figure}

CBR has been proposed as a useful model for AI systems to exhibit human-like reasoning and learning (\cite{craw2018case}). Early work by \cite{kolodner1992introduction} outlines processes like case retrieval, adaptation, and retention that could enable automated CBR. \cite{aamodt1994case} provides a survey outlining foundational issues, key method variations, and different system approaches. Research has since explored techniques like introspective reasoning to improve system knowledge (\cite{leake2010general}), evidence-driven retrieval to bridge gaps between finding and reusing cases (\cite{sizov2015evidence}), and meta-CBR for systems to self-improve through self-understanding (\cite{murdock2008meta}). \cite{craw2018case} argue CBR is well-suited for domains without well-defined theories, as its memory of experiential cases can enable understanding of problem contexts and clustering. Figures~\ref{fig:f1} and~\ref{fig:f2} show the core CBR elements and processes presented in 1992 and 2018. 

CBR's application also raises important questions around value alignment, an active area of research focused on ensuring AI behaves according to human values (\cite{Gabriel_2020_valuealignment, hendrycks2020aligning, chen2023case}). Recent work on value alignment for AI, such as Reinforcement Learning from Human Feedback (RLHF) (\cite{stiennon2020rlhf, ouyang2022rlhf}) and ``Constitutional AI'' applying the abstract principles of harmlessness, honesty, and helpfulness to AI systems (\cite{bai2022constitutional}) has pursued a unitary model of ethics reflecting commonsense human values. However, critics argue this could impose hegemonic perspectives, homogenizing diverse viewpoints (\cite{parrot}). They advocate for more pluralistic representation, noting limitations of prevailing preference-based utilitarian ethics (\cite{tasioulas2022artificial}) and misalignment with demographic groups (\cite{santurkar2023whose, durmus2023anthropic}). 

Alternative techniques have been proposed, such as optimizing text generation for arbitrary rewards (\cite{lu2022quark}), recursive reward modelling from human feedback (\cite{wu2023fine}), and the large-scale ValuePrism dataset connecting pluralistic values to situations (\cite{sorensen2023value}). These methods are meaningful yet do not adequately convey the nuances of individual moral perspectives. We envision that an evolving case base would enable adapting recommendations and explanations to users' specific values, balancing common ground with individual differences. Thus, CBR could inform AI alignment research to achieve ongoing ``reflective equilibrium'' between general principles and particular judgments, similar to how common law incrementally evolves. 

\subsection{Future Work}
We primarily described our process in a linear fashion, but through future work, we strive to make it iterative. New queries received by an AI system may not have closely related precedents in the case repository and will need to be added to the repository as a case. We see this iterative case creation and expansion process as a way of instilling expert and public input as an indispensable ingredient in aligning AI systems. Some experts in our workshops expressed concern over AI's potential automation of high-skill labor and expertise, but with our iterative process, our stance is that it is crucial to engage experts and laypeople alike to contribute towards and continually update knowledge AI may draw upon. We thus invite the community to consider interface affordances and incentive structures for iterative case repository assembly.

We initially ran our process with 20 legal experts, but are starting to branch out to mental health and medical experts. Even though our approach is meant to be domain-agnostic, domain-specific requirements of case repositories may still arise---we look forward to surfacing them to better understand the challenges and needs of diverse communities as they assemble their own case repositories. Additionally, our approach is also designed to be model-agnostic. That is, once a case repository is created, it is a resource that can be made available to a variety of LLMs. We can then test the case repository for performance and consistency across models to both improve our approach and learn more about different models' abilities to adopt CBR.

Finally, case law is primarily practiced in Western legal systems. Future workshops should involve experts familiar with non-Western legal frameworks to discuss the benefits and shortcomings of a CBR-based approach.

\section{Conclusion}

In this paper, we present a process for enabling CBR to complement existing constitutional approaches to AI alignment. Our process integrates knowledge of domain experts, the public, and LLMs to assemble a set of value-encoded precedents in the form of a case repository to inform AI behavior based on a community's existing values. More broadly, we see cases as an enlightening medium for reasoning about morality, given its prior applications in law and philosophy, and by extension, morality in AI. Our efforts are an early attempt at working with this new medium for AI alignment. As interdisciplinary conversations are critical to moving the needle in this space, we invite researchers and practitioners in diverse fields including---but not limited to---AI, philosophy, psychology, cognitive science, and law, to contribute to continuing discussions to refine, critique, and iterate on our proposed ideas.

\begin{ack}
This project was supported by OpenAI's ``Democratic Inputs to AI'' grant program. We thank Teddy Lee, Tyna Eloundou, Aviv Ovadya, members of the grant recipient cohort, and others at OpenAI for insightful discussions and support. 
\end{ack}



\bibliographystyle{plainnat}
\bibliography{refs}


\end{document}